\documentclass[]{llncs}
\usepackage{color}
\usepackage{times}
\usepackage{epsfig}
\usepackage{graphicx}
\usepackage{amsmath}
\usepackage{amssymb}
\usepackage{graphicx}
\usepackage{subcaption}
\usepackage{algorithm}
\usepackage{algpseudocode}
\usepackage[marginal]{footmisc}
\usepackage{mathtools}
\usepackage{enumerate}
\usepackage{paralist}
\usepackage{bm}
\usepackage{multirow}
\usepackage{array}
\usepackage{color}
\usepackage{tabularx,booktabs}
\usepackage{threeparttable}

\newcommand{\cP}{\mathcal{P}}

\newcommand{\etal}{{\it et. al.}}
\newcommand{\ie}{{\it i. e.}}
\newcommand{\eg}{{\it e. g.}}

\usepackage[width=122mm,left=12mm,paperwidth=146mm,height=193mm,top=12mm,paperheight=217mm]{geometry}
\begin{document}
\pagestyle{headings}
\mainmatter
\def\ECCV18SubNumber{}  

\title{Removing out-of-focus blur from a single image}


\author{Guodong Xu, Chaoqiang Liu, Hui Ji}
\institute{Department of Mathematics, NUS}

\maketitle

\begin{abstract}
Reproducing an all-in-focus image from an image with defocus regions is of practical value in many applications, \eg, digital photography, and robotics. Using the output of some existing defocus map estimator, existing approaches first segment a de-focused image into multiple regions blurred by Gaussian kernels with different variance each, and then de-blur each region using the corresponding Gaussian kernel.  In this paper, we proposed a blind deconvolution method specifically designed for removing defocus blurring from an image, by providing effective solutions to two critical problems: 1) suppressing the artifacts caused by segmentation error by introducing an additional variable regularized by weighted $\ell_0$-norm; and 2) more accurate defocus kernel estimation using non-parametric symmetry and low-rank based constraints on the kernel. The experiments on real datasets showed the advantages of the proposed method over existing ones, thanks to the effective treatments of the two important issues mentioned above during deconvolution.
\end{abstract}

\section{Introduction}\label{sec:introduction}
\begin{figure}
		\centering
		\begin{subfigure}[t]{0.45\textwidth}
			\centering
			\includegraphics[width=1\linewidth]{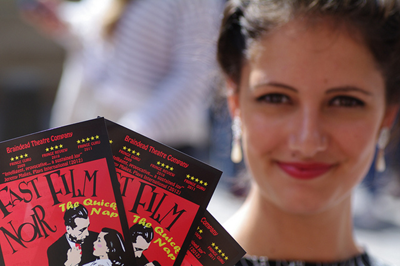}
			\caption{input}
			\label{fig:example:a}
		\end{subfigure}
		\hspace{\fill}
		\begin{subfigure}[t]{0.45\textwidth}
			\centering
			\includegraphics[width=1\linewidth]{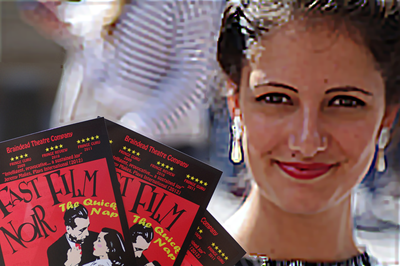}
			\caption{output}
			\label{fig:example:b}
		\end{subfigure}
		\caption{\small (a) input: an image with defocus regions; (b) output: an all-in-focus image.}
		\label{fig:example}
	\end{figure}

The sharpness of an object in a photograph taken by a conventional camera is determined by its scene distance to the focal plane of the camera. The best sharpness is obtained when an object is exactly on the focal plane, \ie, the object is \emph{in-focus}. When an object is away from the focal plane, it will appear blurry. The further is an object away from the focal plane, the more blurry it appears in the image. Such a phenomenon is called \emph{defocus} or \emph{out-of-focus}. When taking pictures of a scene with multiple objects with different depths, the camera with limited depth of field may only focused on one object, and the other part of the image is out-of-focus. See Fig.~\ref{fig:example:a} for an illustration. 

This paper aims at developing an effective method to generate an all-in-focus image from an input by recovering its out-of-focus regions. See Fig.~\ref{fig:example} for an illustration. Such a method is of great practical value to  many applications in machine vision, \eg, photo refocusing, camera systems for robotics, and many others.

\vskip 3pt

\noindent {\bf Issues in existing methods.}\quad
Several approaches have been proposed to produce an all-in-focus image from an image with defocused regions, see \eg~\cite{dai2009removing,shen2012spatially,tung2017multiple,chan2011single}.
Out-of-focus blurring is usually spatially varying since the blurring amount of individual pixel is determined by its scene depth. The existing solution to such a spatially varying deconvolution problem is first segmenting the input image into several regions with approximately uniform blur amount, and then  running blind deconvolution on each region.

\begin{figure}
	\centering
	\begin{subfigure}[t]{0.24\textwidth}
		\centering
		\includegraphics[width=1\linewidth]{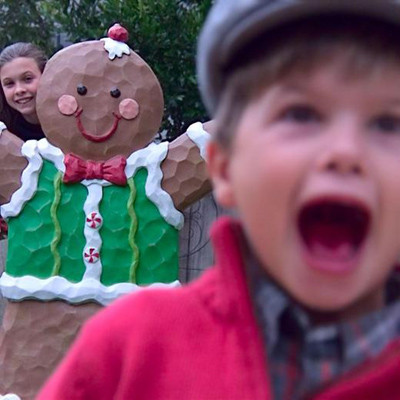}
		\caption{}
	\end{subfigure}
	\hspace{\fill}
		\begin{subfigure}[t]{0.24\textwidth}
		\centering
		\includegraphics[width=1\linewidth]{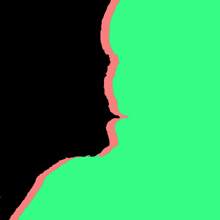}
		\caption{}
	\end{subfigure}
	\hspace{\fill}
			\begin{subfigure}[t]{0.24\textwidth}
		\centering
		\includegraphics[width=1\linewidth]{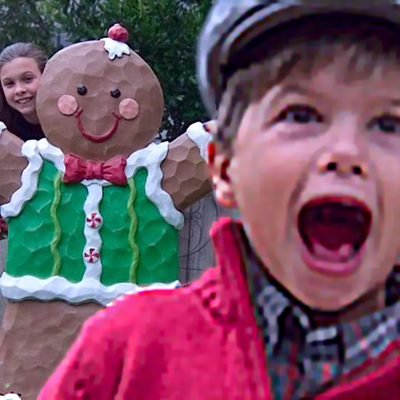}
		\caption{}
	\end{subfigure}
	\hspace{\fill}
	\begin{subfigure}[t]{0.24\textwidth}
		\centering
		\includegraphics[width=1\linewidth]{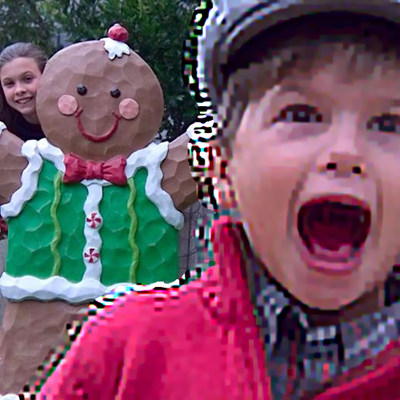}
		\caption{}
	\end{subfigure}
	\caption{\small (a)  input image; (b) accurate segmentation (green part) and in-accurate segmentation with erroneous pixel assignment (indicated in red part);  (c) deblurred result using accurate segmentation; (d) deblurred result  using in-accurate segmentation.}
	\label{fig:example_1}
\end{figure}

In recent years, many effective defocus map estimators have been proposed to estimate the defocus degree of all image pixels; see \eg~\cite{elder1998local,zhuo2011defocus,tang2013defocus,shi2015just,shi2015break,Xu2017}. One important application of defocus map estimator is image segmentation which  segments the image into multiple image regions based on defocus degree; see  \cite{shi2015just,shi2015break,Xu2017} for more details. The segmentation accuracy of these defocus map estimator is satisfactory on tested datasets, but contains errors.  It is well-known that the deconvolution is quite sensitive to the outliers and boundary error. In other words, standard deconvolution methods will have noticeable artifacts in the output  when deblurring an image region that contains image pixels corresponding different blurring degrees. See Fig.~\ref{fig:example_1} for an illustration\footnote{In Section~\ref{sec:introduction}, non-blind image deconvolution is done by standard TV-based regularization method.}

\begin{figure}[!htbp]
	\centering
 \begin{tabular}{ll}
		\raisebox{-0.6\totalheight}{\includegraphics[width=0.4\linewidth]{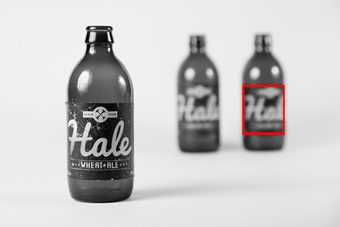}} \quad  &
	\begin{tabular}[t]{ccccc}
					\includegraphics[width=.1\linewidth]{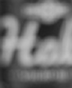}&  \includegraphics[width=.1\linewidth]{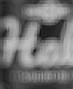}& \includegraphics[width=.1\linewidth]{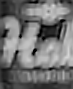}&
					\includegraphics[width=.1\linewidth]{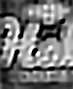}&
					\includegraphics[width=.1\linewidth]{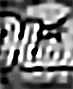}\\
										(a) input & (b) $\sigma=1$& (c)  $\sigma=2$&  (d) $\sigma=3$& (e) $\sigma=4$\\
			\includegraphics[width=.1\linewidth]{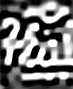}  &   
		 \includegraphics[width=.1\linewidth]{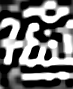}  &
		 \includegraphics[width=.1\linewidth]{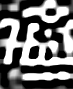}&
		 \includegraphics[width=.1\linewidth]{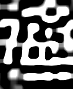}&
		 \includegraphics[width=.1\linewidth]{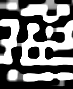}\\
		 (f) $\sigma=5$ & (g) $\sigma=5.5$& (h)  $\sigma=6$&  (i) $\sigma=6.5$ & (j) $\sigma=7$
	\end{tabular}
\end{tabular}
	\caption{\small Illustration of the approximation accuracy of Gaussian kernel to true defocus blur kernel. Image on the left: a real image with defocus regions. (a): blurry image patch  from (a);
	(b)--(j): the results deblurred using  Gaussian kernels with different s.t.d. $\sigma$.}
\label{fig:ill_badgaussian}
\end{figure}

In addition, recovering a high-quality sharp image from its blurred observation  requires the estimation of kernel to be very accurate. Most existing approaches of recovering defocused images use  Gaussian kernel to model defocus kernel, whose approximation accuracy might be adequate for defocus segmentation, but not for the purpose of deblurring. See Fig.~\ref{fig:ill_badgaussian} for an illustration which shows the limitation when modeling defocus kernel by Gaussian function.

In summary, provided the segmentation derived from the defocus map using an existing estimator, an effective  blind deconvolution method  defocus blurring requires good solutions to the following  two questions.
\begin{enumerate}[P1:]
	\item How to improve the robustness of deconvolution to segmentation error.
	\item How to estimate  defocus-type blur kernel with better accuracy than Gaussian.
\end{enumerate}

\vskip 3pt
\noindent {\bf Our contribution.}\quad
In this paper we proposed a blind deconvolution method for defocus blurring that provides effective solutions to the two questions raised above. 
\begin{enumerate}
	\item  A novel regularization for estimating non-parametric form of
	defocus blur kernel.
	\item A new optimization model for deconvolution with an $\ell_0$-norm regularized  term that suppressing artifacts caused by segmentation error.
\end{enumerate}
The experiments show that the proposed optimization approach noticeable outperformed the existing methods in real datasets.

\section{Related work}\label{sec:relatedwork}
Blind image deblurring is a challenging yet important problem in computer vision, particularly owing to the missing information of blurring processing. In recent years, there has been an abundant literature on blind deblurring but with the focus on motion blur.

Motion blur is quite different from out-of-focus blur in terms of characteristics of blur kernel. Motion blur kernel is determined by the relative motion between camera and scene, which roughly can be modeled by a function with curve-like support with strong orientation. Out-of-focus blur kernel is usually a smooth function within a region with strong isotropy. Owing to space limitation, we only mention a few 
results on blind motion deblurring. For blind motion deblurring, both spatially-invariant and spatially-varying cases have been addressed. For removing spatially-invariant motion blur, most existing ones are based on $\ell_p$-norm relating regularization, see \eg,  \cite{fergus2006removing,xu2010two,shan2008high,cai2009blind,Pan2014}. There are also many methods for removing spatially-varying motion blur, see \eg,  \cite{cho_iccv2011,Tai2011,Ji2012,hirschfast,Gupta10,ji2012two}.

Defocus blurring is often spatially-varying invariant, as  the defocus amount of an image pixel is determined by its corresponding scene depth.  The so-called \emph{defocus map}, i.e. the defocus degrees of all image pixels,  is of great interest in computer vision. For example, it is used for  the segmentation of in-focus foreground and defocus background in \cite{shi2015just,shi2015break,Xu2017}.
Defocus amount estimation usually is done on image edges, and then propagated to other image pixels done by matting Laplacian method~\cite{levin2008closed} or inverse diffusion Bae \etal~\cite{bae2007defocus}.   Elder and Zucker~\cite{elder1998local} modeled defocus around an edge as a convolution of a step function with a Gaussian kernel. The s.t.d. of the Gaussian kernel is used for measuring defocus amount. Using the same model as~\cite{elder1998local}, Zhuo and Sim~\cite{zhuo2011defocus} proposed to estimate the blur amount using the ratio of gradient magnitudes between the input image and a re-blurred image convoluted by a Gaussian kernel. Tang \etal~\cite{tang2013defocus} utilizes spectrum contrast to estimate defocus amount at edge locations. Shi \etal~\cite{shi2015just,shi2015break} proposed a method based on the sparse representation over a dictionary learned or handcrafted from a set of images with different contents. Xu~\etal~\cite{Xu2017} proposed a method using the rank of image patches to estimate defocus amount of edge pixels.

Most existing works on removing defocus blurring from images rely on the defocus map provided by some defocus amount estimator discussed in the paragraph above. With a defocus map in hand, the image is segmented into in-focus regions and defocus regions either by thresholding defocus map \cite{zhuo2011defocus,Xu2017}) or running a graph-cut algorithm on defocus map (\eg in~\cite{shi2015just,shi2015break}).  The main differences among the existing defocus blurring methods lie in how to blindly deconvolve defocus image region, which includes blur kernel estimation and non-blind deconvolution.

For deblurred defocus  image regions,  Dai~\etal~\cite{dai2009removing} estimated the  kernel directly from some defocus amount estimator discussed above. Then, the non-blind deconvolution on defocus region is done by using a weighted $\ell_2$-norm based total variation (TV) regularization. Shen \etal \cite{shen2012spatially} modeled a defocus blurred image edge as a step function convoluted by a Gaussian kernel, whose s.t.d was estimated from the ratio between the first derivative extrema and the range of the input image in a local window, followed by guided filtering~\cite{he2013guided}. Then, the defocus image region is recovered by using an $\ell_2$-$\ell_1$ norm based  TV regularization method. Zhang~\etal~\cite{zhang2012single} first segmented the input image into defocus region and in-focus region by thresholding defocus map from some defocus amount estimator. For defocus region, they predict a sharp image region by enhancing the image edge using the Gaussian kernel with s.t.d. estimated by the method \cite{elder1998local}). Then, the blur kernel and clear image are estimated by solving a blind deconvolution problem with specific regularization on the sharp image. Chan \etal's \cite{chan2011single} method combined the idea of Dai~\etal~\cite{dai2009removing} method and and that of Zhang~\etal's~\cite{zhang2012single} method.


\section{Main body}
This paper consider the case where  the scene is composed of multiple depth-ordered layers. Each layer is the convolution between a sharp image part and a defocus blur kernel. In other words, an image with out-of-focus blurring can be modeled as follows.
\begin{equation}
f = \sum_{i=0}^{L}  \alpha_i\odot (k_i \otimes u_i)+ n,
\end{equation}
where $\odot$ denotes the element-wise  product operation, $\otimes$
denotes the convolution operation, and  for the $i$-th layer, $\alpha_i$ denote the binary region matrix whose entries are $1$ for the pixels within the region and $0$ otherwise;
$u_i$ denotes the corresponding sharp image part; $k_i$ denotes the  PSF associated with the $i$-th layer; $n$ denotes the image noise. The output, denoted by $\overline{f}$, will be  an all-in-focus image of the form:
\vskip -5pt
$$
\overline{f}=\sum_{i=0}^L \alpha_i\odot u_i.
$$

In this paper, $\{\alpha_i\}$ is estimated by running the segmentation on the defocus map of the image using some existing one, \eg, Xu \etal's method~\cite{Xu2017}. It is noted that we do not assume that segmentation is  perfect, and one of our focuses is to address such error in the proposed deconvolution method.

\subsection{Optimization model}\label{sec:deblur}
For $i$-th layer, the goal of blind deblurring is to estimate both $k_i$ and $u_i$, given the input:
\begin{equation}
\label{eqn:deblur}
\alpha_i\odot (k_i\otimes u_i)=f_i-\alpha_i\odot n,
\end{equation}
where $f_i=\alpha_i\odot f$.
We take a regularization based approach to solve \eqref{eqn:deblur}. In other words, we simultaneously estimate both the sharp image part $u_i$ and the defocus kernel $k_i$ by solving the following optimization problem:
\vskip -5pt
\begin{equation}
\label{eqn:opm}
\min_{k_i,u_i,c_i}\frac 1 2\|\alpha_i\odot (k_i\otimes u_i)- f_i-c_i\|_F^2+\lambda_1\Phi(u_i)+\lambda_2\Psi(k_i)+\lambda_3\Theta(c_i),\quad
\mbox{s.t.}\quad k_i\in \Omega,
\end{equation}
where $\Phi(u),\Psi(k)$ denote the regularizations on sharp image part and blur kernel, $\Omega$ denotes the feasible set for defocus blur kernel, $c_i$ is a variable for addressing segmentation errors which we will introduce later, and $\Theta(c_i)$ is its regularizations. For image regularization $\Phi(\cdot)$, we adopt the existing widely used $\ell_1$-norm relating regularization:
$$
\Phi(u_i)=\|Wu_i\|_1,
$$
where $W$ could be either difference operator or other operator such as wavelet transform. The spline wavelet transform \cite{Daub03} is used in our implementation.

\subsubsection{Regularization on defocus kernel $\Psi$ and its feasible set $\Omega$.}
Most existing regularization either use a parametric form of Gaussian function, or adopt one from blind motion deblurring to estimate non-parametric form of kernel.  Gaussian kernel provides a rough approximation to practical defocus blur kernel, which is not accurate for the purpose of deblurring, as shown in Fig.~\ref{fig:ill_badgaussian}. The regularizations for motion blur kernels is not suitable for defocus blur, as these two have very different optical characteristics.

\begin{figure}
	\centering
	\subcaptionbox{rank=1}{\includegraphics[width=.15\linewidth]{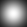}}
	\subcaptionbox{rank=8}{\includegraphics[width=.15\linewidth]{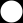}}
	\subcaptionbox{rank=7}{\includegraphics[width=.15\linewidth]{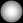}}
	\caption{\small Rank of defocus blur kernels  with tight bounding box. (a)--(c): three types of parametric forms of defocus kernel proposed in existing literatures:  Gaussian kernel with size $27\times 27$ and $\sigma=9$; pillbox kernel with size $23\times 23$ and diameter $23$~\cite{shi2015break}; Gaussian pupil with size $23\times 23$, the diameter of the disk-like pupil is $23$ and the the s.t.d of the Gaussian function is $9$.}
	\label{fig:rank_kernel}
\end{figure}

\begin{figure}
	\centering
	\subcaptionbox{}{\includegraphics[height=0.22\linewidth]{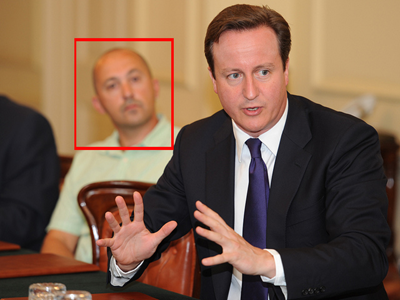}}
	\subcaptionbox{}{\includegraphics[height=0.22\linewidth]{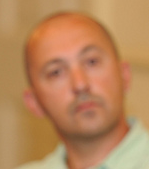}}
	\subcaptionbox{}{\includegraphics[height=0.22\linewidth]{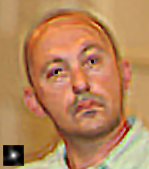}}
	\subcaptionbox{}{\includegraphics[height=0.22\linewidth]{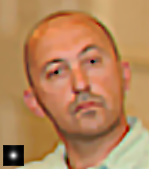}}
	\caption{\small illustration of performance impact of low-rank based constraint. (a) input image; (b) blurred image region; (c) and (d) are deblurred results by model~\eqref{eqn:model:final} without and with low-rank constraint on kernel respectively. The estimated blur kernels are shown in left bottom, their size are both $31\times 31$, while their ranks are $18$ and $2$ respectively.}
	\label{fig:test_lrcon}
\end{figure}

In this paper, we proposed two constraints on practical defocus blur kernel. The first is low rank based constraint. It is observed that all parametric forms of defocus blur kernel proposed in  existing literatures are of low-rank in matrix form. For example, the most often seen Gaussian kernel can be expressed as 
 $$G=gg^\top\in \mathbb{R}^{m\times m},$$
 where $g\in \mathbb{R}$ is a 1D Gaussian filter. Clearly the rank of $G$ is $1$. Similarly, other types of parametric forms of defocus kernel, e.g., pillbox kernel and Gaussian pupil~\cite{goodman2008introduction},
 also see a low rank structure of its matrix form.   See Fig~\ref{fig:rank_kernel} for an illustration of the matrix ranks of parametric defocus kernel presented in existing literatures. The observation from our empirical experiments comes to the same conclusion. Therefore, we proposed the following low-rank based constraint for regularizing defocus blur kernel:
$$
\mathrm{rank}(k)\leq r_0,
$$
where the constant $r_0$ is the
predefined rank threshold\footnote{The value $r_0$ is  adaptive to the kernel $k$ by being set to the number of singular values of $k$ no less than
$\frac{1}{30}$ of its largest singular value}. See Fig.~\ref{fig:test_lrcon} for an illustration of performance impact of low-rank based constraint to deblurring.

%
%

Another important character of defocus blur kernel that differs from motion-blur kernel is its strong symmetry. The imaging principle of the lens tells us that  the defocus kernel usually has strong circular symmetry~\cite{goodman2008introduction}, while motion-blur kernels own strong orientation (see \eg ~\cite{Levin09}). Thus, we proposed a symmetry-based feasible set constraint, $k_i\in \mathcal{O}$, for defocus blur kernel. The set $\mathcal{O}$ is a group of matrix defined by
$$
\mathcal{O}=\{M\in \mathbb{R}^{d\times d}: M=(M)^\vdash=(M)^\dashv=(M)^\top=
(M^\top)^\dashv=(M^\top)^\vdash\},$$
where $(\cdot)^\top, (\cdot)^\dashv, (\cdot)^\vdash$ are transpose, row-wise flip,
and column-wise flip operators defined by: for an matrix $M \in
\mathbb{R}^{m\times n}$,
$$
M^\top (i,j)=M(j,i);\quad M^\dashv (i,j)=M(m-i+1,j),\quad 
M^\vdash(i,j)=M(i,n-j+1).
$$

Together with the normalization constraint and non-negative constraint,  we proposed the following  regularization for defocus kernel:
$$
\Psi(k_i)=\|k_i\|_F^2,
$$
and additional constraint:
\begin{equation}
\label{eqn:kernel_constraint}
k_i\in \Omega:=\{k : \sum_r k[r]=1, \quad k[r]\geq 0;
\quad \mathrm{rank}(k)\leq r_0;\quad \mbox{and}\quad k\in \mathcal{O}\}.
\end{equation}
See Fig.~\ref{fig:test_sysmcon} for an illustration of performance impact of symmetry-based constraint.

\subsubsection{Regularization on terms addressing segmentation error $\Theta(\cdot)$.}
Deconvolution is very sensitive to the outliers and boundary errors. Thus, a segmentation with error  will lead to 
noticeable artifacts if not appropriately treated in the deconvolution.
We propose to address it by introducing an additional variable 
in the fidelity term of \eqref{eqn:opm}. 

For $i$-th layer, let $\hat{\alpha}_i$ be  the estimated mask of
the true mask $\alpha_i$. Define $\delta_{\alpha_i}=\alpha_i-\hat{\alpha}_i$. Assume the blurring on $\delta_{\alpha_i}$ is uniform and $k_\delta$ the blur kernel, otherwise it can be cut into smaller peaces, and similar assumption can be applied on each peace. Let $A_{\kappa}$ be the convolution operator with kernel $\kappa$ and $\delta_{A}=A_{k_i}-A_{k_\delta}$, then 
\begin{equation}
\begin{split}
\alpha_i\odot f &= \hat{\alpha}_i\odot(A_{k_i} u_i)+\delta_{\alpha_i}\odot[(A_{k_i}+A_{\delta})u_i]+\alpha_i\odot n,\\
&=\alpha_i\odot(A_{k_i} u_i)+\delta_\alpha\odot (\delta_{A} u_i)+\alpha_i\odot n,\\
\end{split}
\end{equation}
Notice that $\delta_{A}$ is the difference between two low-pass filters corresponding to two kernels: the blur kernel in the regions indicated by $\delta_{\alpha}$ and the blur kernel corresponding to $i$-th layer. Thus, the term $\delta_{A} u_i$ is actually the response of the sharp image $u_i$ to some high-pass filters, difference between two-pass filters, and the term $\delta_{\alpha}\odot(\delta_{A}u_i)$ is sampling  such output. 

The observation leads us to represent $\delta_{\alpha}\odot(\delta_{A}u_i)$ by the variable $c$ in the optimization model \eqref{eqn:opm}. Since the output of an sharp image convolved with a high-pass filters is usually sparse, we propose  a weighted $\ell_0$-norm relating regularization for the term $c_i$:
\begin{equation}
\label{eqn:lambda}
\Theta(c_i)=\|c_i\|_{0,\Lambda}=\sum_{r}\Lambda[r]
|c_i[r]|.
\end{equation}
The weight matrix $\Lambda$ is determined by the residual from some initial guess of $k_i^{(1)}, u_i^{(1)}$ obtained from the first run of the proposed iterative method:
$$
\Lambda[r]=e^{-500(\alpha_i\odot (k_i^{(1)}\otimes u_i^{(1)})-f_i)[r]}.
$$
In other words, the pixels with larger residual error is more likely to be the pixels that should not be included in the segment. See Fig.~\ref{fig:illust_rac} for an illustration of performance impact of the introduction of $c_i$.

In summary, the optimization model proposed for blind defocus deblurring is expressed as follows.
\begin{equation}
\label{eqn:model:final}
\min_{k_i,u_i,c_i}\frac 1 2\|\alpha_i\odot (k_i\otimes u_i) -f_i-c_i\|_F^2+\lambda_1\|Wu_i\|_1+\lambda_2\|k_i\|_F^2+
\lambda_3\|c_i\|_{0,\Lambda}
\end{equation}
subject to 
$$\Omega=\{k: \sum_r k[r]=1, \quad k[r]\geq 0;
\quad \mathrm{rank}(k)\leq r_0;\quad \mbox{and}\quad k\in \mathcal{O}\},$$
where $\lambda_1, \lambda_2,\lambda_3$ are regularization parameters.

\subsection{Numerical algorithm}
For the optimization problem \eqref{eqn:model:final} above, we take an alternating iteration scheme, which alternatively updates the estimation of the kernel $k_i$, the sharp image part $u_i$, and the residual term $c_i$

\vskip 2mm
\noindent{\bf Updating $u_i$, given $k_i$ and $c_i$}. Given an estimation of defocus kernel $k_i$ and the residual term $c_i$, the problem~\eqref{eqn:model:final} is a often seen convex problem with cost function composed by a differentiable term and a $\ell_1$-norm relating convex term:
\begin{equation}
\label{eqn:model:u1}
\min_{u_i}\  \|\alpha_i\odot (k_i\otimes {u_i})-\alpha_i\odot f-c_i\|_F^2 + \lambda_1\|W u_i\|_1.
\end{equation} 
In recent years, many numerical solvers have been proposed to solve such a convex problem, \eg~the alternating direction method of multipliers (ADMM)~\cite{boyd2011distributed,parikh2014proximal}. Due to space limitation, the readers are referred to \cite{boyd2011distributed,parikh2014proximal} for 
more details.

\vskip 2mm
\noindent{\bf Updating $k_i$, given $u_i$ and $c_i$}. Given an estimation of sharp image part $u_i$ and  the term $c_i$, the problem~\eqref{eqn:model:final} is a constrained problem with a quadratic cost function:
\begin{equation}
\label{eqn:model:k1}
\begin{split}
\min_{k_i}\ &\Gamma(k_i)=\min_{k_i}\|\alpha_i\odot (u_i\otimes k_i) -\alpha_i\odot f-c_i\|_F^2 +\lambda_2\|k_i\|_F^2\\
\text{s.t.}\ &\sum_{r}k_i[r] = 1,\ k_i[r]\geq 0, \ \mathrm{rank}(k_i)\leq r_0, \ k_i\in \mathcal{O}.
\end{split}
\end{equation}
To solve~\eqref{eqn:model:k1}, we adopt the projected gradient algorithm (PGA)~\cite{bertsekas1999nonlinear}. Let $g(k)=\frac{\partial \Gamma(k)}{\partial k}$ denote the gradient, and let $\Omega$ denote the feasible set: 
\begin{equation}\label{eqn:Omega}
\Omega = \{k \ : \ k[r]\geq 0, \sum_{r}k[r]=1, \mathrm{rank}(k)\leq r_0,
\  k\in \mathcal{O}\}.
\end{equation}
 Define the projection operator $\cP_{\Omega}$ onto the feasible set~\eqref{eqn:Omega} by
$$
\mathcal{P}_\Omega: k\rightarrow k_r
\rightarrow k_{\mathcal{O}}= (((((k_r^\vdash+k_r)^\dashv+k_r)^\top+k_r)^\vdash+k_r)^\dashv+k_r)/5
\rightarrow P_+(k_{\mathcal{O}}),
$$
where $P_+: x\rightarrow \max\{x,0\}$, and  $k_r$ denotes the rank-$r_0$ approximation to $k$, which can be ontained by only keeping 
 the $r_0$ largest single values in the Singular Value Decomposition of $k$.
See Alg.~\ref{alg:subgradient} for the details of the PGA method. 
\begin{algorithm}
	\caption{The PGA method for solving~\eqref{eqn:model:k1} }
	(i) \  Set initial guess $k_i^{(0)}$, step size $\eta^{(0)}$.\\
	(ii) For $n=0,1,\dots$, perform the following iterations until convergence,
	\[
	\left\{
	\begin{aligned}
	x^{(n+1)}&:=k_i^{(t)}-\eta^{(n)} g(k_i^{(n)});\\
	k_i^{(n+1)}&:=\cP_{\Omega}(x^{(n+1)});\\
	\eta^{(n+1)}& := \eta^{(n)}/(n+1);
	\end{aligned}
	\right.
	\]
	\label{alg:subgradient}
\end{algorithm}

\vskip 2mm
\noindent{\bf Updating $c_i$, given $u_i$ and $k_i$}. Given an estimation of $u_i$ and $k_i$, the problem~\eqref{eqn:model:final} is a often seen convex problem with cost function composed by a differentiable term and a $\ell_0$-norm relating non-convex term:
\[
\label{eqn:model:c1}
\min_{c_i}\  \|\alpha_i\odot (k_i\otimes {u_i})-\alpha_i\odot f-c_i\|_F^2 + \lambda_3\|c_i\|_{0,\Lambda}.
\]
Although problem~\eqref{eqn:model:c1} is non-convex, it has a close-form minimizer:
\[
c_i= \mathcal{T}_h(\alpha_i\odot(k_i\otimes {u_i}- f);\lambda_3\Lambda),
\]
where $\mathcal{T}_h(x;\Lambda)$ is the hard thresholding operator, which keeps $x[r]$  if $|x[r]|>\Lambda[r]$, and set to $0$ otherwise.
See Alg.~\ref{alg:whole} for the outline of the iterative blind defocus deconvolution for the $i$-th layer.
\begin{algorithm}
	\caption{Iterative method for solving~\eqref{eqn:model:final}}
	\label{alg:whole}
	\begin{algorithmic}[1]
		\State{\bf INPUT:}
		input image $f$, defocus region mask $\alpha_i$, iterations $N$.
		\State{\bf OUTPUT:}
		deblurred image part $u_i$ and defocus kernel $k_i$
		\State Initialize defocus kernel $k_i$.
		\For{$n=0,1,2,\cdots,N-1$}
		\State $u_i^{(n+1)}:=\text{ADMM}(k_i^{(n)},c_i,\alpha_i,f)$;
		\State $k_i^{(n+1)}:=\text{PGA}(u_i^{(n+1)},c_i,\alpha_i,f)$;
		\State $c_i:= \mathcal{T}_h(\alpha_i\odot(k_i\otimes {u_i}-f);\lambda_3\Lambda)$;
		\EndFor
		\State Set $u_i:=u_i^{(N)}$ and $k_i:=k_i^{(N)}$.
	\end{algorithmic}
\end{algorithm}

\subsubsection{Summarization of our approach.} 
\begin{figure}
	\centering
	\includegraphics[width=0.9\linewidth]{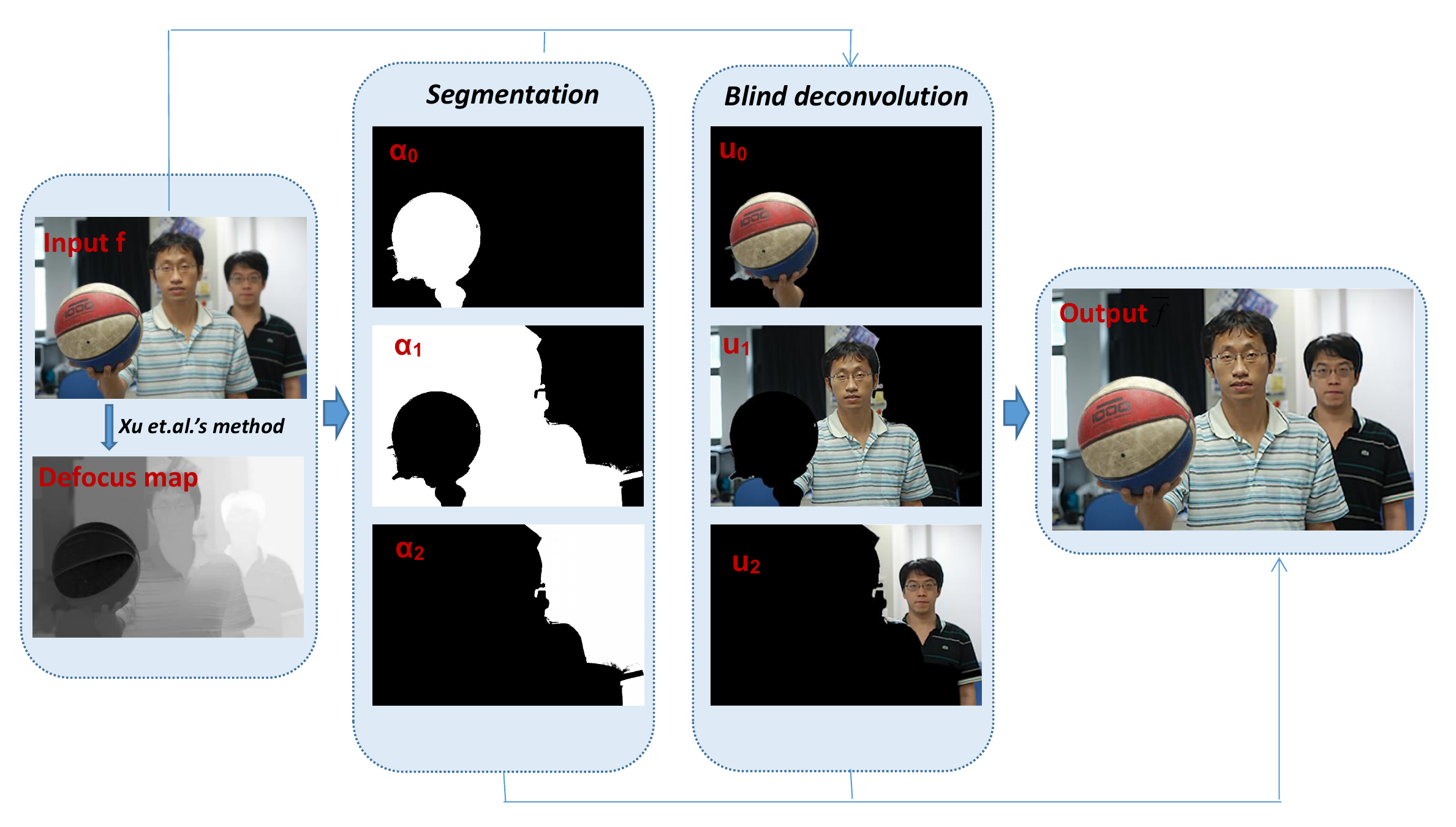}
	\caption{outline of our approach.}
	\label{fig:outline}
\end{figure}

Before ending this section, we would like to summarize the whole scheme of our approach to extract an all-in-focus image from the input image. Given an image  $f$ that contains one or multiple defocus layers, we first estimate its defocus map using some existing defocus map estimator. Then the input image is segmented into an in-focus layer and $L$ out-of-focus layers by thresholding on the defocus map, which leads to the estimation of $\{\alpha_i\}$.
For each out-of-focus layer, Alg.~\ref{alg:whole} is called to generate the underlying sharp image  $u_i$. Finally, the all-in-focus image is generated as
$$
\overline f=\sum_{i=0}^{L}  \alpha_i\odot u_i. 
$$
See Fig.~\ref{fig:outline} for an illustration for the workflow of the proposed approach. 

\section{Experiments}
\subsection{Implementation details}
Give the input image, its defocus region masks $\{\alpha_i\}$ can be estimated using any defocus region segmentation method mentioned in Section~\ref{sec:relatedwork}.
In this paper, we adopt Xu~\etal's method~\cite{Xu2017}.  Interested reader may refer to~\cite{Xu2017} for more details. 

All parameters are set to the same for all experiments, except the parameter $\lambda_2$ which will automatically adjust to different cases.
$$  \mbox{kernel size}=31\times 31, \ N=20,  \lambda_1=0.005, \ \lambda_3=0.01, \
\lambda_2=(-50\|k^{(0)}\|_F+15)\times 10^5,$$
where $k^{(0)}$ denotes some initial estimation of parametric form of disk function to defocus blur kernel. Such initial estimation is done as by following the method proposed in Tung~\etal's~\cite{tung2017multiple}. That is, we  deblur the region using the disk functions with different sizes to pick the one which gives least square error between the blur one and re-blurred result using the estimated kernel. 
The parameter $r_0$ in Alg.~\ref{alg:subgradient} is also adaptive to the input. The rank-$r_0$ approximation of $k$ is done by only keeping the singular values larger than $1/30$ times of its largest singular values in its SVD. 


\subsection{Experiments on simulated data}
\subsubsection{Evaluation of performance of feasible set for kernel.} 
 In this section, the experiments are done to evaluate 
the proposed regularizations on  kernel and segmentation errors.

\begin{figure}
	\centering
	\subcaptionbox{}{\includegraphics[height=0.24\linewidth]{./fig/Deblur_withGaussian/illustration_gvd/blur_whole_gray_1.png}}
	\subcaptionbox{}{\includegraphics[height=0.24\linewidth]{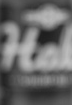}}
	\subcaptionbox{}{\includegraphics[height=0.24\linewidth]{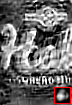}}
	\subcaptionbox{}{\includegraphics[height=0.24\linewidth]{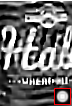}}
	\caption{\small Illustration of performance impact of symmetry-based constraint. (a) real image with defocus regions; (b) blurry image patch from (a); (c) and (d) are deblurred results by model~\eqref{eqn:model:final} without and with symmetric-based constrain respectively, where the estimated blur kernels are shown in right bottom.}
	\label{fig:test_sysmcon}
\end{figure}

 {\bf A. Symmetry-based constraint on kernel}.
For one defocus-blurred image region  of a real image from the dataset\footnote{\url{http://www.cse.cuhk.edu.hk/~leojia/projects/dblurdetect/dataset.html}}, the proposed algorithm is used to deblur the image region twice. One is with symmetry-based constraint,
$k_i\in \mathcal{O}$, and the other drops symmetry-based constraint, while keeping other parts and parameter values the same.
See Fig.~\ref{fig:test_sysmcon} for the comparison. It can be seen that
the method with symmetry-based constraint on kernel outperformed the same one without symmetry-based constraint.

 {\bf B. low-rank based constraint on kernel.}
Similarly, the result from the proposed method with low-rank based constraint on kernel is compared to the one from the proposed method without low-rank based constraint. Consistent with Experiment A, the inclusion of low-rank based constraint improves the performance of the proposed method, as better results are obtained on the test image.

\begin{figure}
	\centering
	\begin{subfigure}[t]{0.19\textwidth}
		\centering
		\includegraphics[width=1\linewidth]{./fig/illus_seg/input.jpg}
		\caption{}
		\label{fig:illust_rac:a}
	\end{subfigure}
	\hfill
	\begin{subfigure}[t]{0.19\textwidth}
		\centering
		\includegraphics[width=1\linewidth]{./fig/illus_seg/mask_3B_b_c.png}
		\caption{}
		\label{fig:illust_rac:b}
	\end{subfigure}
	\hfill
	\begin{subfigure}[t]{0.19\textwidth}
		\centering
		\includegraphics[width=1\linewidth]{./fig/illus_seg/de_wrongmask.png}
		\caption{}
	\end{subfigure}
	\hfill
	\begin{subfigure}[t]{0.19\textwidth}
		\centering
		\includegraphics[width=1\linewidth]{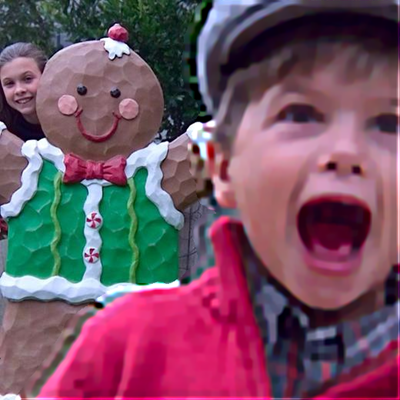}
		\caption{}
	\end{subfigure}
	\hfill
	\begin{subfigure}[t]{0.19\textwidth}
		\centering
		\includegraphics[width=1\linewidth]{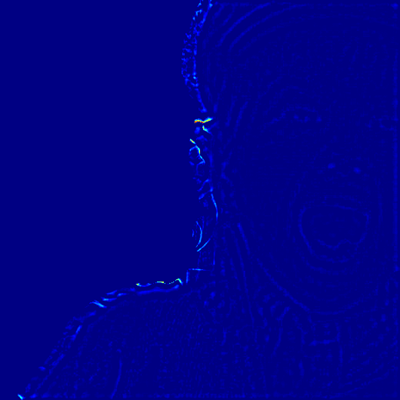}
		\caption{}
	\end{subfigure}
	\caption{\small illustration of robustness of the proposed method. (a) real image with one defocus region; (b) estimated defocus region mask with wrong segmented piece indicated in red; (c) and (d) are the deblurred results of model~\eqref{eqn:model:final} without and with the term $c$ respectively (blur kernel is estimated using perfect segmentation); (f) estimated term $c$.}
	\label{fig:illust_rac}
\end{figure}

{\bf C. Weighted $\ell_0$-norm relating regularization for robustness to segmentation error.} The robustness to segmentation error is important for deblurring an image region. The proposed model introduced a variable $c$ regularized by its weighted $\ell_0$-norm
to gain such robustness. For the illustration, a real image with single out-of-focus region is taken from the dataset~\cite{nejati2015multi}. True segmentation of the defocus region, which is indicated in the green part of Fig.~\ref{fig:illust_rac:b}, is hand-crafted. Then, the segmentation is defined with a small erroneous region indicated in the red part in Fig.~\ref{fig:illust_rac:b}. Then, same as the previous experiments, the region is deblurred by the proposed method 
twice: one with residual term $c$ and the other without residual term $c$. See Fig.~\ref{fig:illust_rac} for the comparison. It can be seen that, the residual term is helpful to suppress ring artifacts caused by segmentation error.

\begin{figure}
	\centering
	\begin{subfigure}[t]{0.24\textwidth}
		\centering
		\includegraphics[width=1\linewidth]{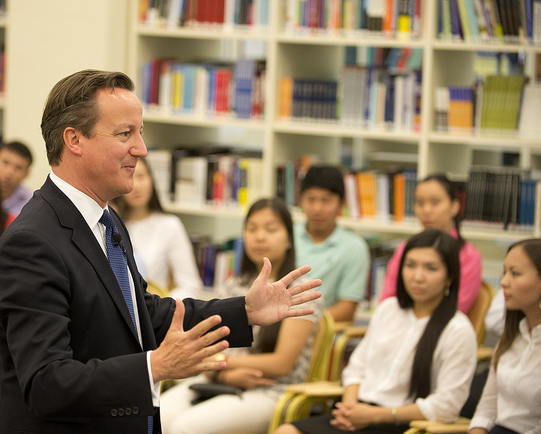}\\
		\includegraphics[height=0.38\linewidth]{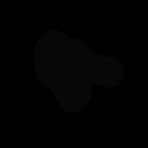}
		\hspace{\fill}
		\includegraphics[height=0.38\linewidth]{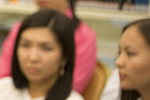}
		\caption{input image}
	\end{subfigure}
	\centering
	\begin{subfigure}[t]{0.24\textwidth}
		\centering
		\includegraphics[width=1\linewidth]{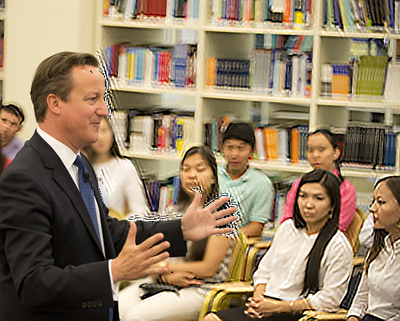}\\
		\includegraphics[height=0.38\linewidth]{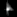}
		\hspace{\fill}
		\includegraphics[height=0.38\linewidth]{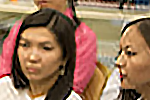}
		\caption{Xu~\etal's~\cite{xu2010two}}
	\end{subfigure}
	\begin{subfigure}[t]{0.24\textwidth}
		\centering
		\includegraphics[width=1\linewidth]{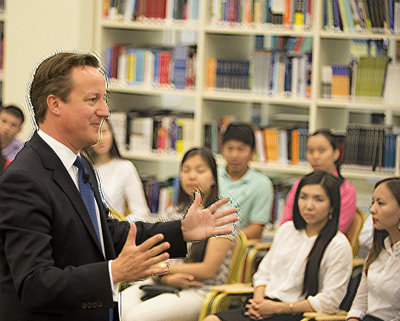}\\
		\includegraphics[height=0.38\linewidth]{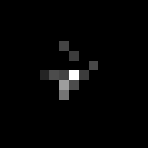}
		\hspace{\fill}
		\includegraphics[height=0.38\linewidth]{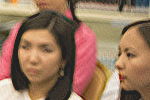}
		\caption{Fergus~\etal's~\cite{fergus2006removing}}
	\end{subfigure}
	\begin{subfigure}[t]{0.24\textwidth}
		\centering
		\includegraphics[width=1\linewidth]{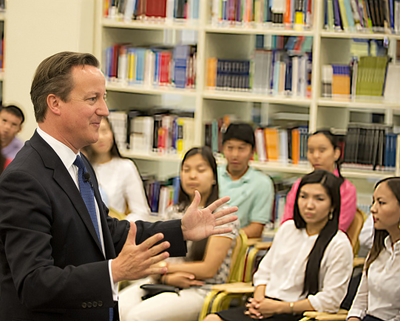}\\
		\includegraphics[height=0.38\linewidth]{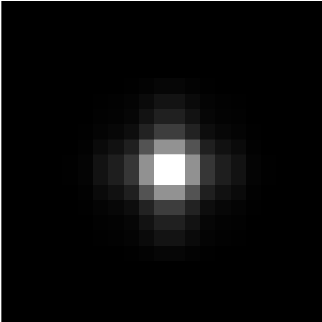}
		\hspace{\fill}
		\includegraphics[height=0.38\linewidth]{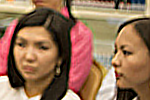}
		\caption{ours}
	\end{subfigure}
	\caption{\small comparison to two existing blind motion deblurring methods.
		The two images in the second row are estimated kernel and zoomed -in regions.}
	\label{fig:def_vs_motion}
\end{figure}

 {\bf D. Comparison to the regularization borrowed from motion deblurring}. This experiment is to show how the specifically designed regularization for defocus kernel is compared to the widely available regularizations on motion blur kernel. See Fig.~\ref{fig:def_vs_motion} for the comparison between the result from the proposed method and the results from two representative blind motion deblurring methods: Fergus~\etal's~\cite{fergus2006removing} and Xu~\etal's~\cite{xu2010two}.  All three methods take the same segmentation as the input.  It can be seen that the proposed one is sharper than the other two with more details, which results from a more accurate estimate of defocus kernel. Indeed, the kernel obtained from the proposed method is very different from that from two blind motion deblurring method.

\subsection{Real data experiments}
In this section, the proposed method is compared to two existing defocus deblurring methods: one is Shen \etal's method ~\cite{shen2012spatially} and the other is Dai \etal's method \cite{dai2009removing} on two real image datasets~\cite{shi2015just,nejati2015multi}. The results from Shen \etal's method are generated from the authors' implementation online. The results from  Dai \etal's method are generated from our implementation of the method  since no code from the authors is available. The method uses the same segmentation method as ours for deblurring. The other related method has no code available online, and an accurate implement is not possible owing to the lack of details and parameter setting.

\begin{figure}
	\centering
	\begin{subfigure}[t]{0.24\textwidth}
		\includegraphics[width=\linewidth]{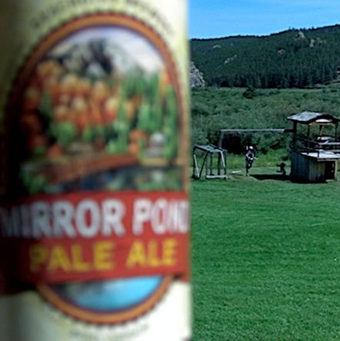}
	\end{subfigure}
	\begin{subfigure}[t]{0.24\textwidth}
		\includegraphics[width=\linewidth]{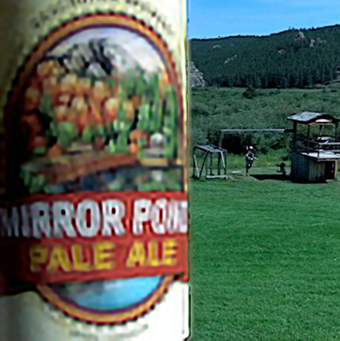}
	\end{subfigure}
	\begin{subfigure}[t]{0.24\textwidth}
		\includegraphics[width=\linewidth]{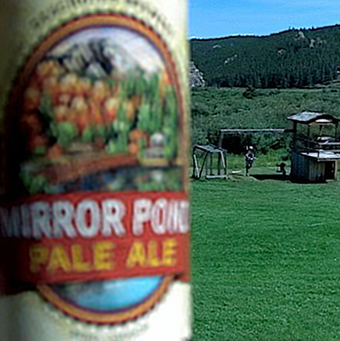}
	\end{subfigure}
	\begin{subfigure}[t]{0.24\textwidth}
		\includegraphics[width=\linewidth]{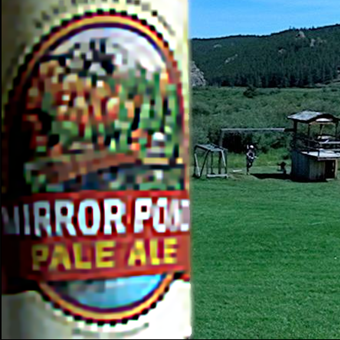}
	\end{subfigure}

\begin{subfigure}[t]{0.24\textwidth}
	\includegraphics[width=1\linewidth]{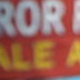}
\end{subfigure}
\begin{subfigure}[t]{0.24\textwidth}
	\includegraphics[width=1\linewidth]{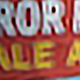}
\end{subfigure}
\begin{subfigure}[t]{0.24\textwidth}
	\includegraphics[width=1\linewidth]{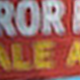}
\end{subfigure}
\begin{subfigure}[t]{0.24\textwidth}
	\includegraphics[width=1\linewidth]{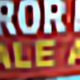}
\end{subfigure}

%
%
%
	
	\centering
	\begin{subfigure}[t]{0.24\textwidth}
		\includegraphics[width=1\linewidth]{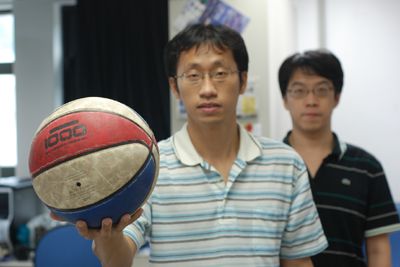}
	\end{subfigure}
	\begin{subfigure}[t]{0.24\textwidth}
		\includegraphics[width=1\linewidth]{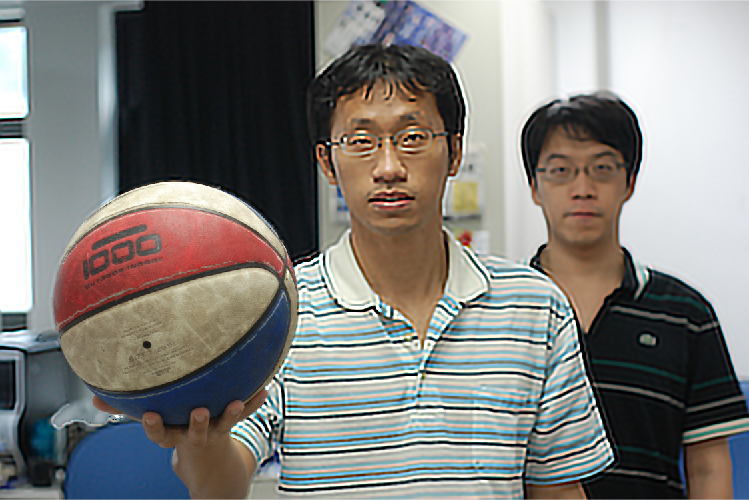}
	\end{subfigure}
	\begin{subfigure}[t]{0.24\textwidth}
		\includegraphics[width=1\linewidth]{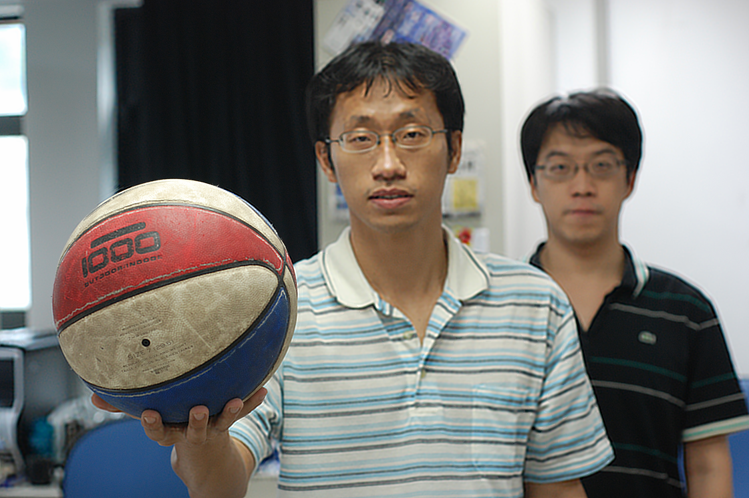}
	\end{subfigure}
	\begin{subfigure}[t]{0.24\textwidth}
		\includegraphics[width=1\linewidth]{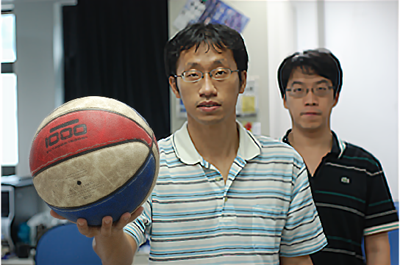}
	\end{subfigure}
	
	\begin{subfigure}[t]{0.24\textwidth}
		\includegraphics[width=\linewidth]{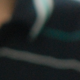}
	\end{subfigure}
	\begin{subfigure}[t]{0.24\textwidth}
		\includegraphics[width=\linewidth]{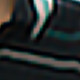}
	\end{subfigure}
	\begin{subfigure}[t]{0.24\textwidth}
		\includegraphics[width=\linewidth]{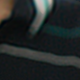}
	\end{subfigure}
	\begin{subfigure}[t]{0.24\textwidth}
		\includegraphics[width=\linewidth]{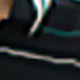}
	\end{subfigure}

	\begin{subfigure}[t]{0.24\textwidth}
		\includegraphics[width=\linewidth]{./fig/org/656.png}
	\end{subfigure}
	\begin{subfigure}[t]{0.24\textwidth}
		\includegraphics[width=\linewidth]{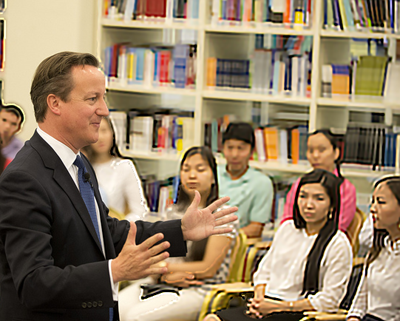}
	\end{subfigure}
	\begin{subfigure}[t]{0.24\textwidth}
		\includegraphics[width=\linewidth]{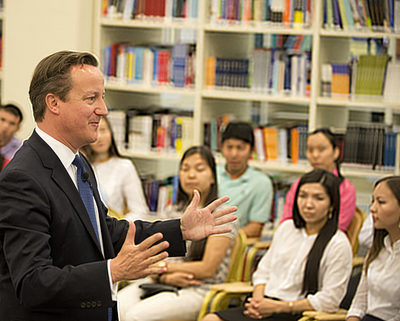}
	\end{subfigure}
	\begin{subfigure}[t]{0.24\textwidth}
		\includegraphics[width=\linewidth]{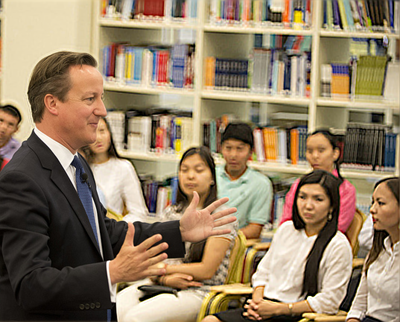}
	\end{subfigure}
	
	\begin{subfigure}[t]{0.24\textwidth}
		\includegraphics[width=\linewidth]{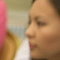}
		\caption{input image}
	\end{subfigure}
	\begin{subfigure}[t]{0.24\textwidth}
		\includegraphics[width=\linewidth]{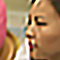}
		\caption{Dai \etal's~\cite{dai2009removing}}
	\end{subfigure}
	\begin{subfigure}[t]{0.24\textwidth}
		\includegraphics[width=\linewidth]{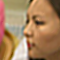}
		\caption{Shen \etal's~\cite{shen2012spatially}}
	\end{subfigure}
	\begin{subfigure}[t]{0.24\textwidth}
		\includegraphics[width=\linewidth]{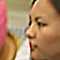}
		\caption{ours}
	\end{subfigure}

	\caption{\small Comparison of deblurred images with both 2-layer
		and 3-layer}
	\label{fig:compare_xuandshen_mult}
\end{figure}

Due to the space limitation, the defocus maps of the images used for segmentation are also included in the supplementary materials. See Fig.~\ref{fig:compare_xuandshen} and Fig.~\ref{fig:compare_xuandshen_mult} for a visual comparison of the results from three methods on four real images.  More experiments on real images can be found in the supplementary materials.

Among these four images,  two  contain one in-focus and one defocus layer, and the other two  contain one in-focus and two different defocus layers. For the images with only one defocus layer, it can be seen that for those defocus layers with large defocus degree, the results from the proposed method show sufficient details and sharper edges in defocus regions. In contrast, the visual quality of the results from Dai~\etal's and Shen~\etal's methods is quite poor.
When the defocus degree is relatively small, our result is comparable with Shen~\etal's and is noticeably better than Dai~\etal's.
For the images with two different defocus layers, it can be seen that both of Dai~\etal's and Shen~\etal's methods fail to recover sharp image edges on the layer with relatively large defocus degree, and Dai~\etal's tends to over deblurred the layer with relatively small defocus degree. It can be seen that our method produced noticeable better results than the two methods for comparison.
%


\section{Conclusions}
In this paper, we proposed an approach to remove out-of-focus blurring from a single image which provided effective solutions to  addressed two key questions in blind defocus deblurring: robustness to segmentation error and effective constraints for estimating practical defocus blur kernel. The experiments on real data sets showed the advantages of the proposed method over existing ones.

\begin{figure}
	\centering
\begin{subfigure}[t]{0.24\textwidth}
	\includegraphics[width=\linewidth]{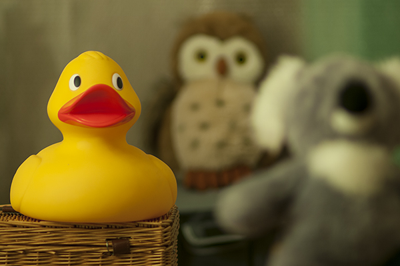}
\end{subfigure}
\begin{subfigure}[t]{0.24\textwidth}
	\includegraphics[width=\linewidth]{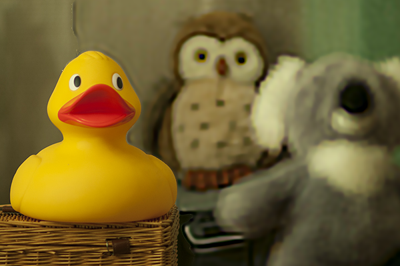}
\end{subfigure}
\begin{subfigure}[t]{0.24\textwidth}
	\includegraphics[width=1\linewidth]{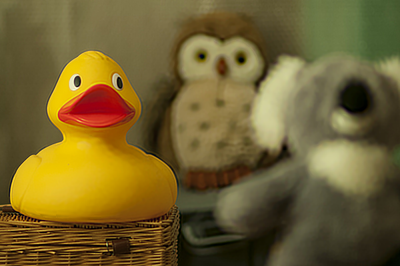}
\end{subfigure}
\begin{subfigure}[t]{0.24\textwidth}
	\includegraphics[width=1\linewidth]{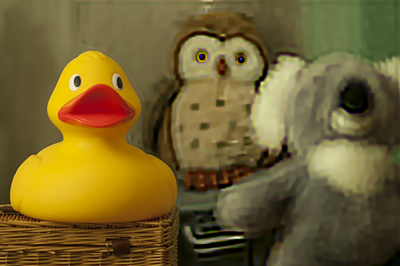}
\end{subfigure}

\begin{subfigure}[t]{0.24\textwidth}
	\includegraphics[width=\linewidth]{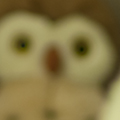}
	\caption{input image}
\end{subfigure}
\begin{subfigure}[t]{0.24\textwidth}
	\includegraphics[width=1\linewidth]{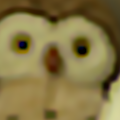}
	\caption{Dai \etal's~\cite{dai2009removing}}
\end{subfigure}
\begin{subfigure}[t]{0.24\textwidth}
	\includegraphics[width=1\linewidth]{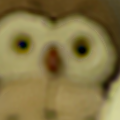}
	\caption{Shen \etal's~\cite{shen2012spatially}}
\end{subfigure}
\begin{subfigure}[t]{0.24\textwidth}
	\includegraphics[width=1\linewidth]{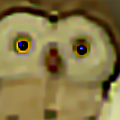}
		\caption{ours}
\end{subfigure}

	\caption{\small Comparison of deblurred images with 3-layer.}
	\label{fig:compare_xuandshen}
\end{figure}

\clearpage

\bibliographystyle{splncs}
\bibliography{BDD_R1_iso}

\end{document}